\documentclass{article} 
\usepackage{iclr2026_conference,times}

\usepackage{amsmath,amsfonts,bm}

\def\eqref#1{equation~\ref{#1}}

\def\1{\bm{1}}

\DeclareMathAlphabet{\mathsfit}{\encodingdefault}{\sfdefault}{m}{sl}
\SetMathAlphabet{\mathsfit}{bold}{\encodingdefault}{\sfdefault}{bx}{n}

\usepackage[utf8]{inputenc} 
\usepackage[T1]{fontenc}    
\usepackage{hyperref}       
\usepackage{url}            
\usepackage{booktabs}       
\usepackage{amsfonts}       
\usepackage{amsmath}
\usepackage{multirow}
\usepackage{float}
\usepackage{subcaption}
\usepackage{wrapfig}
\usepackage{comment}

\definecolor{ForestGreen}{RGB}{34,139,34}

\newcommand{\boldsmallforestgreen}[1]{{\textbf{\tiny\textcolor{ForestGreen}{#1}}}}

\usepackage{todonotes}

\title{Beyond Speedup - Utilizing KV Cache for Sampling and Reasoning}

\author{
Zeyu Xing$^1$, Xing Li$^2$, Hui-Ling Zhen$^2$, Mingxuan Yuan$^2$, Sinno Jialin Pan$^1$ \\
$^1$Department of Computer Science and Engineering, The Chinese University of Hong Kong \\
$^2$Huawei Technologies Co., Ltd. \\
\texttt{zeyuxing@link.cuhk.edu.hk} \\
\texttt{\{li.xing2, zhenhuiling2, yuan.mingxuan\}@huawei.com} \\
\texttt{sinnopan@cuhk.edu.hk}
}

\iclrfinalcopy 

\begin{document}

\maketitle

\begin{abstract}
KV caches, typically used only to speed up autoregressive decoding, encode contextual information that can be reused for downstream tasks at no extra cost. We propose treating the KV cache as a lightweight representation, eliminating the need to recompute or store full hidden states. Despite being weaker than dedicated embeddings, KV-derived representations are shown to be sufficient for two key applications: \textbf{(i) Chain-of-Embedding}, where they achieve competitive or superior performance on Llama-3.1-8B-Instruct and Qwen2-7B-Instruct; and \textbf{(ii) Fast/Slow Thinking Switching}, where they enable adaptive reasoning on Qwen3-8B and DeepSeek-R1-Distil-Qwen-14B, reducing token generation by up to $5.7\times$ with minimal accuracy loss. Our findings establish KV caches as a free, effective substrate for sampling and reasoning, opening new directions for representation reuse in LLM inference. \textbf{Our code is available at} \url{https://github.com/cmd2001/ICLR2026_KV-Embedding}.
\end{abstract}

\section{Introduction}

Large language models (LLMs) rely on key-value (KV) cache to accelerate autoregressive decoding by reusing past attention states, avoiding costly recomputation. This makes the KV cache indispensable for low-latency inference in production systems like vLLM~\citep{Kwon2023Efficient}. However, its role is typically confined to this speedup. Beyond acceleration, the KV cache is seldom viewed as a reusable representation—with the notable exception of cache steering, a technique that modifies the cache's initial state to guide generation~\citep{belitsky2025kvsteer}.

\begin{wrapfigure}{r}{0.45\textwidth}
    \centering\vspace{-5mm}
    \includegraphics[width=0.9\linewidth]{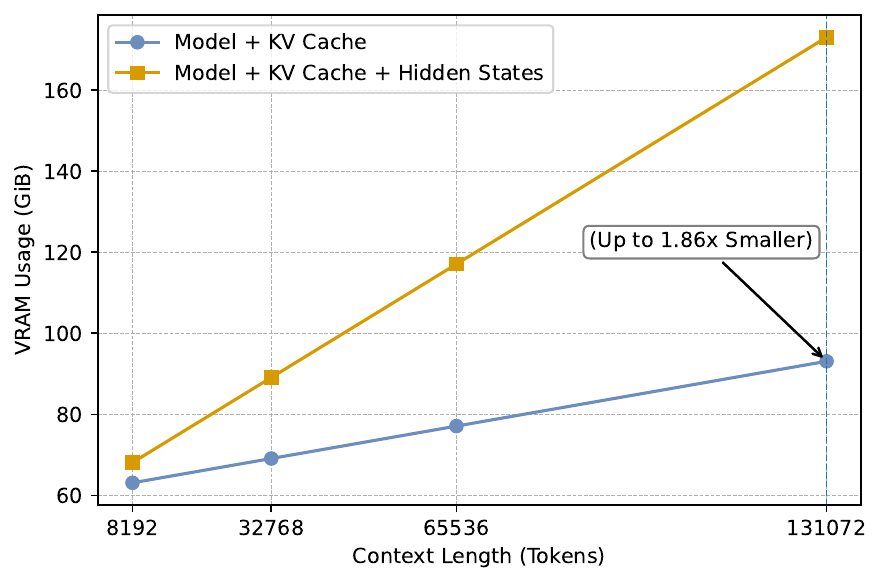}
    \vspace{-2mm}
    \caption{VRAM usage vs.\ context length for Qwen3-32B~\citep{yang2025qwen3}, 
             comparing Model+KV Cache vs.\ Model+KV Cache+Hidden States.}
    \label{fig:kv_cache_hidden_state_comparison}
    \vspace{-5mm}
\end{wrapfigure}

While the KV cache has been mostly confined to acceleration~\citep{li2025kvtuner, yangattentionpredictor}, hidden states have been widely exploited for \emph{self-evaluation}~\citep{yuan2024coe, chen2025inside, beigi2024internalinspector, chen2025probing_right} and for \emph{adaptive reasoning and control}~\citep{zhang2025asrr, wang2025pats, wang2025dots}. These methods, however, rely on storing full hidden states, which is costly in both memory and compute.

In this work, we investigate a simple but powerful question: \textbf{Can the KV cache do more than just accelerate decoding?} Since the KV cache is already computed and stored during inference, using it for downstream tasks incurs \emph{no additional cost}. This is a major advantage over storing full hidden states, which is prohibitively expensive in terms of memory. As shown in Figure~\ref{fig:kv_cache_hidden_state_comparison}, the KV cache offers a significantly more compact and practical alternative for typical decoder-only models.

Though the KV cache is not explicitly trained as a general-purpose embedding—its sole objective is to support next-token prediction—we find it nonetheless encodes rich contextual information suitable for various downstream tasks. We explore this potential through two applications:
\begin{itemize}
    \item \textbf{Chain-of-Embedding:} We repurpose the KV cache as a lightweight and readily available embedding. In experiments on Chain of Embedding (CoE)~\citep{yuan2024coe}—a method for selecting optimal reasoning paths without external information—we show that KV caches achieve classification performance comparable to or even surpassing that of using the model's hidden states.
    \item \textbf{Fast/Slow Thinking Switch:} We leverage the KV cache to implement an adaptive switching mechanism between fast, low-compute reasoning and slower, deliberate reasoning. By reusing KV cache, this approach achieves substantial efficiency gains with minimal performance loss.
\end{itemize}

Our contributions are fourfold:
\begin{enumerate}
    \item We present the first systematic study of KV caches as reusable task representations, showing they can be repurposed at near-zero computational cost. In particular, we propose simple but effective aggregation techniques that make KV caches directly usable as embeddings.  
    \item Despite not being designed as general-purpose embeddings, we find that KV cache representations when processed with the proposed aggregation strategies, are competitively effective on certain classification tasks.
    \item We propose \textbf{KV-CoE}, a variant of Chain-of-Embedding that reuses the KV cache  already stored during decoding. KV-CoE achieves self-evaluation without extra activation storage, offering nearly zero memory overhead and seamless integration into existing inference frameworks.
    \item We introduce \textbf{KVClassifier}, a fast/slow auto-thinking framework that reuses KV caches for adaptive reasoning with minimal overhead.
\end{enumerate}

Our results suggest that KV caches are a versatile and low-cost foundation for sampling and reasoning, moving beyond their traditional role as a mere acceleration component to become a core resource for effective and efficient LLM-based inference.

\section{Related Work}

\paragraph{Hidden–state self-evaluation.}
A growing line of work shows that internal activations encode reliable signals about answer correctness and hallucination risk.
\citet{yuan2024coe} propose \emph{Chain-of-Embedding} (CoE), which models the trajectory of layerwise hidden states during inference and derives output-free correctness scores from the geometry of this path.
\citet{chen2025inside} (INSIDE) introduce \emph{EigenScore}, computed from the eigenspectrum of hidden-state covariance, to assess semantic (in)consistency and detect hallucinations.
\citet{beigi2024internalinspector} train a contrastive probe on \emph{internal states} (attention, MLP activations) to produce well-calibrated confidence estimates across NLU/NLG tasks.
\citet{chen2025probing_right} further probes hidden states of reasoning models to predict whether a generated answer will be correct.
All of these methods operate directly on hidden states or logits. {\bf Our study}, by contrast, investigates whether \emph{the KV cache alone}—which is already present at inference—suffices to support the same families of subtasks.

\paragraph{Adaptive fast/slow reasoning and dynamic control.}
To mitigate overthinking on easy inputs and underthinking on hard ones, recent work explores \emph{adaptive} reasoning depth~\citep{zeyu2025large}.
\citet{zhang2025asrr} quantify upper bounds of long- vs.\ no-thinking modes and propose \emph{Adaptive Self-Recovery Reasoning} (ASRR), adding accuracy-aware length rewards to reduce unnecessary reasoning while allowing implicit recovery.
{PATS}~\citep{wang2025pats} performs \emph{process-level} switching via process reward models with beam search, enabling step-wise fast/slow adaptation with bad-step penalties.
{DOTS}~\citep{wang2025dots} views reasoning as a search over atomic actions and learn to select dynamic trajectories.
These approaches typically require explicit chain-of-thought generation, external reward models, or re-decoding.
{\bf Our contribution} is orthogonal: we show that pooled \emph{KV-cache} features can drive both one-shot (classification-style) and in-generation (generative-style) switching through simple control tokens, without storing hidden states or altering model architecture.

\paragraph{KV-cache interventions.}
While our work treats the KV cache as a read-only representation for evaluation and control, a concurrent line of research shows that it can also serve as a \emph{control interface}.
\citet{belitsky2025kvsteer} introduce \emph{KV Cache Steering}, a one-shot intervention that adds layer-wise steering vectors—derived from contrastive CoT vs.\ non-CoT prompts—to the key and value tensors after prefill, reliably inducing longer and more structured reasoning in small Language Models.
Compared with activation steering, this method offers improved stability and negligible runtime overhead.
{\bf Our approach} is complementary: instead of modifying the cache, we \emph{pool} it to derive difficulty-aware signals that gate slow reasoning.
\section{Background}

\subsection{Transformer, Hidden States, and KV Cache}

Decoder-only transformers are the architectural foundation of modern large language models (LLMs)~\citep{vaswani2017attention,brown2020gpt3}. During autoregressive generation, each transformer layer processes a new token to produce a contextual \emph{hidden state}. A computational bottleneck arises because standard attention requires recomputing over all previous tokens at each step, resulting in $\mathcal{O}(T^2)$ complexity per step, where $T$ is the sequence length. To mitigate this, the {key--value (KV) cache} \citep{dao2022flashattention} stores the attention keys and values for all past tokens at every layer. This allows the model to compute keys and values only for the new token and attend to the cached history, reducing the complexity to $\mathcal{O}(T)$ per step and enabling efficient long-sequence generation.

Formally, for layer $l$, we store
\[
\text{KVCache}^{(l)} = \{K^{(l)}_{1:T}, V^{(l)}_{1:T}\},
\]
where $K^{(l)}_{1:T}, V^{(l)}_{1:T} \in \mathbb{R}^{T \times H \times d_{\text{head}}}$ are the stacked key and value tensors across all attention heads $H$. The hidden state at step $t$ is compuated via
\[
h^{(l)}_t = \text{Attention}\!\left(Q^{(l)}_t, K^{(l)}_{1:t}, V^{(l)}_{1:t}\right).
\]
Since $K^{(l)}_{1:t-1}$ and $V^{(l)}_{1:t-1}$ are already cached, only $K^{(l)}_t$ and $V^{(l)}_t$ need to be computed online.

\subsection{Modern LLM Frameworks and KV Cache Management}
State-of-the-art LLM serving frameworks carefully manage KV caches to achieve high throughput, low latency, and efficient GPU memory utilization.

Modern LLM frameworks treat the KV cache as a first-class resource. vLLM \citep{Kwon2023Efficient} introduces \emph{PagedAttention}, virtualizing the KV cache with a paging mechanism akin to CPU virtual memory to enable dynamic allocation, low fragmentation, and high-throughput serving under heavy concurrency. In contrast, Ollama \citep{ollama2024} focuses on lightweight, developer-friendly deployment, managing KV caches at the session level to efficiently reuse context across multi-turn interactions. Together, these systems illustrate how both production-scale and interactive inference workloads rely on persistent, reusable KV caches as a core abstraction.

Across these frameworks, the KV cache is managed as a first-class resource with explicit strategies for allocation, eviction, and reuse. This observation motivates our central claim:

``\emph{Since the KV cache is an unavoidable byproduct of efficient inference, repurposing it for downstream tasks adds virtually no overhead}.''
\section{Observation}
\subsection{Can KV Caches Serve as an Embedding Source}

The hidden states and attention projections stored in KV caches encode contextualized token representations, making them natural candidates for use as embeddings. While recent work has explored leveraging intermediate representations from LLMs as task-specific embeddings \citep{liu2024llmembed}, we specifically investigate aggregating KV cache vectors into sentence-level representations.  

To evaluate their quality as an embedding source, we construct embeddings by concatenating keys and values at every layer, then averaging across token positions, attention heads, and layers before applying $\ell_2$ normalization. We benchmark these KV-derived embeddings on the Massive Text Embedding Benchmark (MTEB) \citep{muennighoff2023mteb} against a strong, dedicated embedding model (gemini-embedding-001).  

\begin{table}[htbp]
\centering
\resizebox{0.9\linewidth}{!}{
\begin{tabular}{lcc}
\toprule
\textbf{Dataset} & \textbf{Llama-3.1-8B-Instruct KV Cache} & \textbf{Gemini-Embedding-001} \\
\midrule
AmazonCounterfactualClassification & 0.3530 & 0.8820 \\
DBpediaClassification              & 0.5937 & 0.9476 \\
FinancialPhrasebankClassification  & 0.6254 & 0.8864 \\
TweetTopicSingleClassification     & 0.3714 & 0.7111 \\
\bottomrule
\end{tabular}
}
\vspace{-1mm}
\caption{Performance of KV cache-based embeddings vs.\ a dedicated embedding model on selected MTEB classification tasks. Despite being significantly weaker than trained embeddings, KV-derived embeddings still capture meaningful semantics. Appendix~\ref{appendix:mteb} further reports hidden-state and chance baselines under the same evaluation pipeline.}
\label{tab:kv_mteb}
\vspace{-3mm}
\end{table}
As shown in Table~\ref{tab:kv_mteb}, KV-derived embeddings significantly underperform their dedicated counterpart across all datasets, confirming they are \emph{not perfect general-purpose embeddings}. This gap stems from three factors: (i) KV representations are optimized for causal language modeling, not contrastive learning, leading to poor isotropy; (ii) they are inherently token- and position-specific, requiring heuristic pooling for sentence-level use; and (iii) their projection into a lower-dimensional head space ($d_{\text{head}} \ll d_{\text{model}}$) reduces their discriminative power.  

Despite these limitations, the results show that KV caches encode substantial semantic information—enough to be competitive on certain classification tasks. This finding motivates our exploration of reusing the KV cache for \emph{Chain-of-Embedding} and \emph{Fast/Slow Thinking Switch}, where global embedding quality is less critical than local, relative separability between candidates.

\subsection{Why KV Caches Are Sufficient for Chain-of-Embedding and Fast/Slow Thinking Switch?}

KV caches are poor \emph{general-purpose} embeddings: they are trained for next-token prediction, are position/context dependent, and the space is often anisotropic. Still, they are {\bf sufficient} for our two uses—{\em Chain-of-Embedding} (CoE) and {\em Fast/Slow Thinking Switch}—because both rely on \emph{local, task-conditioned} comparisons rather than globally calibrated semantics.

\paragraph{Local (restricted-set) adequacy.}
General embedding learning targets global separation across a broad label/instance space. Here we only need correct \emph{relative ordering} within a small candidate set $\mathcal{C}$ (e.g., a small label space or a few candidate continuations). Concretely, for a decision rule $f$ and score gap
\[
\gamma(x)=f_{y_i}(x)-f_{y_j}(x),
\]
we only require $\min_{y\in\mathcal{C}}\gamma(y)>0$, not a globally well-structured embedding space. This is why KV embeddings can work well on \emph{certain classification-like} cases with limited labels/candidates, despite weak MTEB-style semantic similarity.

\paragraph{CoE is not classification.}
CoE extracts a \emph{path embedding} along the reasoning trajectory and estimates the chance of correction from local geometry (e.g., angle/step length). The requirement is therefore weaker than global separability: KV embeddings only need to preserve these \emph{local trajectory cues} consistently over short ranges.

\paragraph{Task conditioning + efficiency.}
The pooled KV embedding $e=\text{Pool}(K,V)$ is conditioned on input and instruction, $e=g(x,\iota)$, which already biases the representation toward the current task. Finally, reusing KV caches is essentially free compared to storing hidden states ($C_{\text{hidden}}\!\gg\! C_{\text{KV}}\!\approx\!0$), so in memory/latency-sensitive regimes the utility favors KV reuse even if accuracy is slightly lower.

\paragraph{Scope and limitations.}
We do \emph{not} claim KV caches yield universally strong embeddings. They are unsuitable when a task needs globally comparable semantics across diverse queries (e.g., broad retrieval/similarity). Our claim is limited to (i) restricted candidate sets and (ii) CoE-style local trajectory geometry.

\section{Chain of Embedding with KV Cache}

\subsection{Background}
LLMs exhibit emergent reasoning capabilities, though their internal decision processes remain largely opaque. To address this, \citet{yuan2024coe} introduce \textit{Chain-of-Embedding} (CoE), a method that probes the model's latent space by tracking the evolution of sentence-level representations across layers. Formally, for an LLM $\mathcal{M}$ with $L$ layers, let $h_l^{(t)}$ denote the hidden representation of token $t$ at layer $l$. The sentence-level representation at layer $l$ is obtained by by averaging over the sequence length $T$:
\begin{equation}
s_l = \frac{1}{T} \sum_{t=1}^{T} h_l^{(t)}, \quad l=0,1,\dots,L.
\end{equation}
The CoE trajectory is then defined as the sequence of these layer-wise representations:
\begin{equation}
\mathrm{CoE} = \{ s_0, s_1, \dots, s_L \}.
\end{equation}
CoE characterizes this trajectory by measuring both magnitude and directional changes of embeddings between consecutive layers:
\begin{equation}
\Delta r_l = \| s_{l+1} - s_l \|_2, \mbox{ and }\;
\Delta \theta_l = \arccos \!\left( \frac{s_{l+1} \cdot s_l}{\|s_{l+1}\| \|s_l\|} \right).
\end{equation}
These features are aggregated into self-evaluation scores. For instance, the real-space combination (CoE-R) is
\begin{equation}
\mathrm{CoE\text{-}R} = \frac{1}{L-1} \sum_{l=0}^{L-1} 
\left( \alpha \Delta r_l + \beta \Delta \theta_l \right),
\end{equation}
where $\alpha,\beta$ are weighting coefficients. A more robust complex-space variant (CoE-C) treats each $(\Delta r_l, \Delta \theta_l)$ pair as a complex number $z_l = \Delta r_l + i\Delta\theta_l$ and computes the magnitude of their average:
\begin{equation}
\mathrm{CoE\text{-}C} = \left| \frac{1}{L-1} \sum_{l=0}^{L-1} z_l \right|.
\end{equation}
CoE has demonstrated strong discriminative power in distinguishing correct from incorrect model generations, achieving state-of-the-art performance on self-evaluation benchmarks.

\subsection{Methodology}

Our key innovation is to adapt the CoE framework to use the KV cache, eliminating its primary computational overhead. While vanilla CoE constructs trajectories from hidden states $h_l^{(t)}$
 , requiring expensive activation storage or re-computation, we instead leverage the key-value tensors $K^{(l,t)},V^{(l,t)}$ that are already maintained by autoregressive decoders. This modification preserves the CoE analytical framework while rendering it virtually cost-free.

\paragraph{Embedding Construction.}
For each token $t$ and layer $l$, we start with the cached key-value tensors $K^{(l,t)},V^{(l,t)} \in \mathbb{R}^{H \times d}$. We flatten the head and key/value dimensions and average across layers to produce a compact per-token embedding:
\begin{equation}
e_t = \frac{1}{L} \sum_{l=1}^L \mathrm{flatten}\!\big(K^{(l,t)},V^{(l,t)}\big) \in \mathbb{R}^{H\cdot d}.
\end{equation}
The resulting token-wise trajectory is defined as:
\begin{equation}
\mathrm{KV\text{-}CoE} = \{ e_1, e_2, \dots, e_T \},
\end{equation}
which directly parallels the structure of vanilla CoE but operates along the token dimension.

\paragraph{Trajectory Characterization.}
We characterize this trajectory using the established CoE metrics, simply substituting the token index $t$ for the layer index $l$:
\begin{align}
&\Delta r_t = \| e_{t+1} - e_t \|_2, \mbox{ and }\;
\Delta \theta_t = \arccos\!\Big(\tfrac{e_{t+1}\cdot e_t}{\|e_{t+1}\|_2 \,\|e_t\|_2}\Big), \\
&\mathrm{KV\text{-}CoE\text{-}R} = \tfrac{1}{T-1}\sum_{t=1}^{T-1}\big(\alpha\,\Delta r_t + \beta\,\Delta \theta_t\big), \mbox{ and }\;
\mathrm{KV\text{-}CoE\text{-}C} = \Big|\tfrac{1}{T-1}\sum_{t=1}^{T-1} (\Delta r_t + i\,\Delta \theta_t)\Big|.
\end{align}
These formulations maintain the analytical rigor of CoE-R and CoE-C with minimal conceptual alteration.

\begin{wrapfigure}{r}{0.4\linewidth}
  \centering\vspace{-5mm}
  \includegraphics[width=0.8\linewidth]{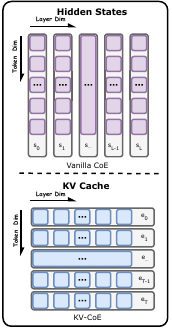}
  \vspace{-0.2cm}
\caption{Comparison between vanilla CoE (top) and \textbf{KV-CoE} (bottom). 
Vanilla CoE aggregates hidden states across the token dimension to construct 
a representation for each layer, whereas KV-CoE aggregates KV Cache 
across the layer dimension to construct a representation for each token.}
  \label{fig:coe}
  \vspace{-15mm}
\end{wrapfigure}

\textbf{Contributions and Advantages.}  As illustrated in Figure \ref{fig:coe}, which compares vanilla CoE and our \textbf{KV-CoE}, our method offers two main advantages:
\begin{enumerate}
    \item \textbf{No extra activation cost.}  
    Since the KV cache is already computed and stored during standard autoregressive decoding, reusing it for trajectory analysis incurs virtually no additional activation cost. The required reductions are computationally negligible compared to a full forward pass, resulting in $\Delta M \approx 0$ extra memory and minimal FLOPs.
    
    \item \textbf{Deployment-friendly.}  
    The approach works directly with standard inference stacks (e.g., \texttt{past\_key\_values} in Transformers or vLLM). It requires no architectural changes, re-forwarding, or activation hooks, making it immediately deployable in production LLM serving systems.
\end{enumerate}

\subsection{Experimental Results}
\label{sec:experiments}

We evaluate \textbf{KV-CoE} on two reasoning benchmarks from the original CoE work: 
MATH~\citep{hendrycks2021math} for multi-step arithmetic and TheoremQA~\citep{chen2023theoremqa} for theorem proving. 
Experiments are conducted on two popular instruction-tuned models: Llama-3.1-8B-Instruct~\citep{grattafiori2024llama3} and Qwen2-7B-Instruct~\citep{yang2025qwen2}.

We construct embeddings directly from the KV cache by extracting value vectors at every layer, concatenating across attention heads, and averaging over layers to obtain one embedding per token, all without storing hidden states. This reuse of cached information introduces negligible VRAM overhead and incurs minimal FLOPs consumption compared to vanilla CoE.

\begin{table}[htbp]
\centering
\resizebox{0.8\textwidth}{!}{%
\begin{tabular}{llcccc}
\toprule
\textbf{Model} & \textbf{Method} &
\multicolumn{2}{c}{\textbf{MATH}} &
\multicolumn{2}{c}{\textbf{TheoremQA}} \\
 & & \textbf{AUROC} $\uparrow$ & \textbf{FPR95} $\downarrow$
   & \textbf{AUROC} $\uparrow$ & \textbf{FPR95} $\downarrow$ \\
\midrule
\multirow{7}{*}{\shortstack{Llama-3.1-8B-\\Instruct}}
  & MaxProb       & 59.16 & 87.96 & 45.41 & 98.60 \\
  & PPL           & 60.82 & 86.42 & 46.45 & 97.82 \\
  & Entropy       & 62.74 & 84.14 & 47.37 & 97.82 \\  \cmidrule(lr){2-6}
  & CoE-R$^\dagger$(Llama3-8B) & 72.54 & 75.61 & 63.12 & 89.83 \\
  & CoE-C$^\dagger$(Llama3-8B) & 73.08 & 79.60 & 55.85 & 90.14 \\ \cmidrule(lr){2-6}
  & \textbf{KV-CoE-R (ours)} & \textbf{64.36} & \textbf{63.82} & 74.74 & 62.93 \\
  & KV-CoE-C (ours)          & 64.13 & 67.42 & \textbf{74.93} & \textbf{62.46} \\
\cmidrule(lr){1-6}
\multirow{7}{*}{\shortstack{Qwen2-7B-\\Instruct}}
  & MaxProb       & 12.40 & 99.34 & 4.92 & 99.87 \\
  & PPL           & 12.43 & 99.50 & 5.11 & 100.00 \\
  & Entropy       & 16.19 & 99.42 & 5.28 & 99.87 \\ \cmidrule(lr){2-6}
  & CoE-R         & 75.75 & 65.95 & 66.68 & 85.84 \\
  & CoE-C         & 76.68 & 64.48 & 62.70 & 87.42 \\ \cmidrule(lr){2-6}
  & KV-CoE-R (ours)      & 76.92 & 49.83 & \textbf{88.87} & \textbf{54.30} \\
  & \textbf{KV-CoE-C (ours)} & \textbf{84.12} & \textbf{44.82} & 83.27 & 58.35 \\
\bottomrule
\end{tabular}}
\vspace{-0.2cm}
\caption{{Self-evaluation results on reasoning tasks.} 
KV-CoE consistently improves AUROC and reduces FPR95 relative to MaxProb, PPL, and Entropy. 
Bold indicates the best value per model–dataset pair except CoE baselines.
CoE-R and CoE-C results are taken from the original CoE paper~\citep{yuan2024coe}. Appendix~\ref{appendix:coe} clarifies CoE usage (confidence estimation, not reranking) and motivates the token-centric KV-CoE design via layer-wise ablations.
{$^\dagger$These baseline results are reported on Llama3-8B-Instruct, 
while our experiments use the updated Llama3.1-8B-Instruct, so the numbers may not perfectly align.}}
\label{tab:kvcoe_results}
\vspace{-0.7cm}
\end{table}

\paragraph{Analysis.}  
As shown in Table~\ref{tab:kvcoe_results}, KV-CoE substantially outperforms baselines such as MaxProb, PPL, and Entropy on both MATH and TheoremQA. This demonstrates that the Chain-of-Embedding approach retains its strong discriminative power even when using KV cache-derived trajectories instead of hidden states. The token-level evolution captured by the KV cache provides a rich signal for identifying correct reasoning paths, especially in multi-step problems, all while adding negligible overhead since the cache is inherently available.

\section{Fast/Slow Thinking Switching with KV Cache}

\subsection{Background}

\begin{wrapfigure}{r}{0.485\linewidth} 
  \centering
  \vspace{-6mm} 
  \includegraphics[width=\linewidth]{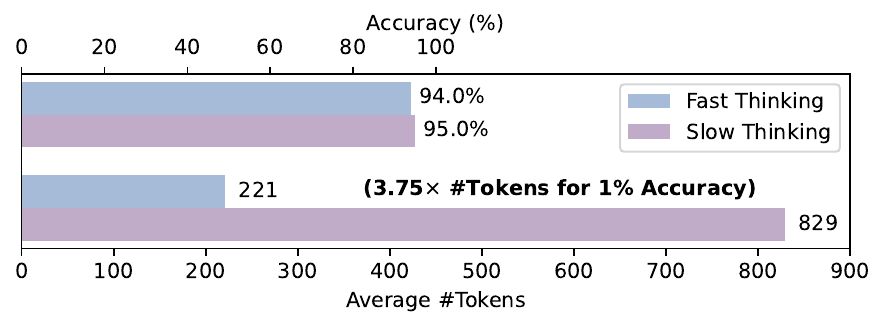}
  \caption{Comparison of efficiency and effectiveness of fast vs.\ slow thinking on GSM8K using Qwen3-32B. Slow thinking achieves slightly higher accuracy but at a much higher token cost.}
  \label{fig:gsm8k_comparison}
  \vspace{-0.2cm}
\end{wrapfigure}

Large Reasoning Models (LRMs) can operate in two modes: \emph{fast thinking}, which produces short, direct answers, and \emph{slow thinking}, which generates detailed, explicit step-by-step reasoning chains.~\citep{yao2023tree,lightman2023lets}. Although slow thinking enhances reliability on complex tasks, it incurs substantial computational overhead by producing significantly more tokens. For example, on GSM8K~\citep{cobbe2021gsm8k}, Qwen3-32B~\citep{yang2025qwen3} slow thinking yields a marginal improvement in accuracy (0.95 vs.\ 0.94) while generating nearly four times the tokens, drastically increasing latency and cost as shown in Figure~\ref{fig:gsm8k_comparison}. This inefficiency motivates \textbf{adaptive reasoning}, where slow thinking is triggered selectively based on problem difficulty.

\subsection{Methodology}

We propose a method for adaptive reasoning that selects between fast and slow thinking on a per-instance basis to minimize unnecessary computation while maintaining accuracy. Our approach leverages the \textbf{KV cache} from the prompt encoding phase to make this decision, eliminating the need for additional forward passes.

\paragraph{Key Idea.}  
Instead of predicting a binary ``slow or fast'' mode, we estimate a continuous {difficulty score} $d \in [0, 100]$ from the pooled KV cache representation:
\[
d = f_\theta\!\left(\text{Pool}\!\left(KV^{(1:L)}_{1:T}\right)\right),
\]
where $\text{Pool}(\cdot)$ aggregates keys and values across layers, heads, and token positions via mean pooling, and $f_\theta(\cdot)$ is a lightweight MLP classifier. This score determines whether to engage slow thinking.  

\paragraph{Switching Mechanism.}  
We control the reasoning mode by injecting special control tokens (\texttt{<think>} and \texttt{</think>}) during decoding:
\begin{itemize}
    \item \textbf{Initial Decision:}  
    Before generation starts, $d$ is compared to a predefined threshold $\tau$:  
    \begin{itemize}
        \item If $d > \tau$, prepend \texttt{<think>} to trigger slow thinking.
        \item Otherwise, proceed with fast thinking.
    \end{itemize}
    
    \item \textbf{Dynamic Adjustment During Decoding:}  
    During generation, the difficulty score is recomputed from the updated KV cache at predefined checkpoints:
    \begin{itemize}
        \item If $d < \tau_{\text{fast}}$ during slow thinking, append \texttt{</think>} to switch back to fast mode.
        \item If $d > \tau_{\text{slow}}$ during fast mode, inject \texttt{<think>} to re-engage slow thinking and continue decoding with step-by-step reasoning.
    \end{itemize}
\end{itemize}

\begin{figure}[t!]
  \centering
  \includegraphics[width=0.9\linewidth]{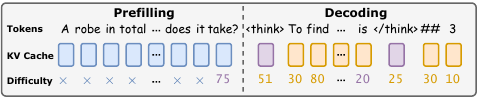}
  \caption{KVClassifier: special tokens are dynamically inserted to perform thinking-mode switching based on KV-derived difficulty scores.}
  \label{fig:kv_clasifier}
  \vspace{-0.5cm}
\end{figure}

This approach enables a fine-grained, difficulty-aware control over reasoning depth. Since the KV cache is already available from prompt encoding, both initial and ongoing difficulty assessments add negligible overhead. This significantly reduces token generation and latency for easy queries while allocating more resources to challenging problems. The overall workflow is illustrated in Figure~\ref{fig:kv_clasifier}.

\paragraph{Training Data Construction.}
To train the difficulty estimator $f_\theta(\cdot)$, we construct supervision signals from public reasoning datasets (training splits of GSM8K~\citep{cobbe2021gsm8k} and MATH ~\citep{hendrycks2021math}).
For each training question, we generate two candidate solutions using the base model: a \emph{fast thinking} response (no chain-of-thought) and a \emph{slow thinking} response (with chain-of-thought). We then extract the final answers and compare them against the ground truth.

Based on the outcomes, we assign a difficulty label $d \in \{0, 25, 75, 100\}$ reflecting the required reasoning depth:
\begin{itemize}
    \item $d=0$ (Very easy): Fast-thinking answer is correct and short ($<128$ tokens).
    \item $d=25$ (Moderate): Fast-thinking answer is correct but long ($\geq 128$ tokens).
    \item $d=75$ (Hard): Fast-thinking answer is wrong, Slow-thinking answer is correct.
    \item $d=100$ (Very hard): Both answers are incorrect.
\end{itemize}

This labeling scheme creates a natural difficulty progression, enabling $f_\theta(\cdot)$ to learn a smooth score that correlates with both correctness and reasoning effort. The token-length criterion distinguishes trivial questions from those needing lengthy outputs without explicit reasoning. The trained estimator provides the continuous difficulty score needed for our adaptive switching mechanism. See Appendix~\ref{appendix:fs-labels-prompts} for the exact label distribution and the prompt templates used to instantiate fast vs.\ slow thinking.

\subsection{Experimental Results}

\paragraph{Setup.}  
We evaluate our KV-cache-based fast/slow thinking mechanism on two reasoning benchmarks: GSM8K evaluation split~\citep{cobbe2021gsm8k} and MATH500~\citep{math5002025}. Our experiments compare two switching strategies: 
\begin{itemize}
    \item \textbf{One-step switch (KV-Classification):} 
    This strategy makes a single decision at generation start based on the predicted difficulty score, committing to either slow or fast thinking for the entire decoding process. It functions as a \emph{classification-style} controller.
    
    \item \textbf{Two-step switch (KV-Generative):} 
    This method performs an initial mode selection and continuously monitors difficulty during decoding. If difficulty drops below $\tau_{\text{fast}}$ during slow thinking, it appends \texttt{</think>} to 
    terminate reasoning early; if difficulty exceeds $\tau_{\text{slow}}$ during fast thinking, it injects \texttt{<think>} to engage slow thinking mid-generation. This implements a \emph{generative-style} controller that dynamically adjusts reasoning depth.
\end{itemize}

We deploy both strategies on two open-weight models: DeepSeek-R1-14B~\citep{guo2025deepseek} and Qwen3-8B~\citep{yang2025qwen3}, evaluating their ability to selectively trigger slow thinking and reduce unnecessary token generation.

We construct representations by concatenating key and value tensors across all heads, summing over selected token positions, and averaging across selected layers without normalization, then feed the result into a two-layer MLP (hidden dimension 512, ReLU activation) for difficulty prediction. This design directly reuses the KV cache available during decoding, introduces negligible VRAM or FLOPs overhead, and functions as a modular component that can be seamlessly integrated into existing inference pipelines without modifications to the base model.

\begin{table}[t!]
\centering
\resizebox{0.75\columnwidth}{!}{
\begin{tabular}{ccll}
\toprule
\textbf{Dataset} & \textbf{Method} & \textbf{DeepSeek-R1-14B} & \textbf{Qwen3-8B} \\
\midrule
\multirow{4}{*}{\textbf{GSM8K}} 
 & Fast Thinking & 0.845 / \;\;218\;\;\;\;\;\;\;\; & 0.904 / \;\;211\;\;\;\;\;\;\;\; \\
 & Reasoning & 0.847 / \;\;432\;\;\;\;\;\;\;\; & 0.933 / 1632\;\;\;\;\;\;\; \\
 & KV-Classification & 0.845 / \;\;218 \;\boldsmallforestgreen{-49.5\%} & 0.914 / \;\;554 \;\boldsmallforestgreen{-66.1\%} \\
 & KV-Generative & 0.835 / \;\;242 \;\boldsmallforestgreen{-44.0\%} & 0.892 / \;\;273 \;\boldsmallforestgreen{-83.3\%} \\
\midrule
\multirow{4}{*}{\textbf{MATH500}} 
 & Fast Thinking & 0.536 / \;\;540\;\;\;\;\;\;\;\;\; & 0.568 / \;\;616\;\;\;\;\;\;\;\; \\
 & Reasoning & 0.590 / 1839\;\;\;\;\;\;\;\; & 0.610 / 4150\;\;\;\;\;\;\; \\
 & KV-Classification & 0.578 / 1506 \boldsmallforestgreen{\;-18.1\%} & 0.604 / 3963 \boldsmallforestgreen{\;\;\;-4.5\%} \\
 & KV-Generative & 0.566 / \;\;657 \boldsmallforestgreen{-64.3\%} & 0.578 / \;\;727 \boldsmallforestgreen{\;-82.5\%} \\
\bottomrule
\end{tabular}}
\caption{Comparison of accuracy and average token usage for fast thinking, full reasoning, and our KV-cache-based switching methods on GSM8K and MATH500. 
For each KV-based method, we report the result from best hyper-parameter configuration identified in Appendix~\ref{appendix:kv_hyperparam}.}
\label{tab:fastslow_results}
\vspace{-0.3cm}
\end{table}

\paragraph{Analysis.}  
As shown in Table~\ref{tab:fastslow_results}, our KV-cache-based switching approach achieves an effective balance between accuracy and efficiency. For instance, on MATH500 using Qwen3-8B, two-step generative switching reduces average token count from 4,150 (full reasoning) to 727 ($5.7\times$ reduction) with only a minimal $3.2\%$ accuracy drop. The one-step classification strategy is more conservative, using more tokens but achieving near-full-reasoning accuracy (0.604 vs.\ 0.610). Similar trends are observed on GSM8K, where KV-cache-based switching maintains high accuracy (up to 0.914) while significantly reducing token consumption compared to full reasoning. These results demonstrate that difficulty scores derived from the KV cache generalize well across tasks and models, enabling efficient and effective adaptive reasoning with negligible overhead.

\section{Conclusion}

This work repurposes the KV cache, moving beyond its conventional role in decoding acceleration to unlock its potential as a versatile, cost-free representation. We demonstrate that although not designed as general-purpose embeddings, KV caches encode rich contextual information that can be effectively utilized for downstream tasks without incurring additional computational overhead. 
Our experiments establish two practical applications: (i) \textbf{Chain-of-Embedding}, where KV-derived embeddings match or surpass the performance of hidden-state embeddings, and (ii) \textbf{Fast/Slow Thinking Switching}, which uses KV-cache-based difficulty scores to enable adaptive reasoning-reducing token usage by up to $5.7\times$ with minimal accuracy loss. These findings position the KV cache as a deployment-friendly substrate for advanced inference techniques, opening new avenues for reusing inference-time artifacts to improve efficiency and controllability in LLMs.

\newpage
\bibliography{reference}

@inproceedings{vaswani2017attention,
 author = {Ashish Vaswani and
Noam Shazeer and
Niki Parmar and
Jakob Uszkoreit and
Llion Jones and
Aidan N. Gomez and
Lukasz Kaiser and
Illia Polosukhin},
 bibsource = {dblp computer science bibliography, https://dblp.org},
 booktitle = {Advances in Neural Information Processing Systems 30: Annual Conference
on Neural Information Processing Systems 2017, December 4-9, 2017,
Long Beach, CA, {USA}},
 editor = {Isabelle Guyon and
Ulrike von Luxburg and
Samy Bengio and
Hanna M. Wallach and
Rob Fergus and
S. V. N. Vishwanathan and
Roman Garnett},
 pages = {5998--6008},
 title = {Attention is All you Need},
 url = {https://proceedings.neurips.cc/paper/2017/hash/3f5ee243547dee91fbd053c1c4a845aa-Abstract.html},
 year = {2017}
}

@inproceedings{brown2020gpt3,
 author = {Tom B. Brown and
Benjamin Mann and
Nick Ryder and
Melanie Subbiah and
Jared Kaplan and
Prafulla Dhariwal and
Arvind Neelakantan and
Pranav Shyam and
Girish Sastry and
Amanda Askell and
Sandhini Agarwal and
Ariel Herbert{-}Voss and
Gretchen Krueger and
Tom Henighan and
Rewon Child and
Aditya Ramesh and
Daniel M. Ziegler and
Jeffrey Wu and
Clemens Winter and
Christopher Hesse and
Mark Chen and
Eric Sigler and
Mateusz Litwin and
Scott Gray and
Benjamin Chess and
Jack Clark and
Christopher Berner and
Sam McCandlish and
Alec Radford and
Ilya Sutskever and
Dario Amodei},
 bibsource = {dblp computer science bibliography, https://dblp.org},
 booktitle = {Advances in Neural Information Processing Systems 33: Annual Conference
on Neural Information Processing Systems 2020, NeurIPS 2020, December
6-12, 2020, virtual},
 editor = {Hugo Larochelle and
Marc'Aurelio Ranzato and
Raia Hadsell and
Maria{-}Florina Balcan and
Hsuan{-}Tien Lin},
 title = {Language Models are Few-Shot Learners},
 url = {https://proceedings.neurips.cc/paper/2020/hash/1457c0d6bfcb4967418bfb8ac142f64a-Abstract.html},
 year = {2020}
}

@inproceedings{dao2022flashattention,
 author = {Tri Dao and
Daniel Y. Fu and
Stefano Ermon and
Atri Rudra and
Christopher R{\'{e}}},
 bibsource = {dblp computer science bibliography, https://dblp.org},
 booktitle = {Advances in Neural Information Processing Systems 35: Annual Conference
on Neural Information Processing Systems 2022, NeurIPS 2022, New Orleans,
LA, USA, November 28 - December 9, 2022},
 editor = {Sanmi Koyejo and
S. Mohamed and
A. Agarwal and
Danielle Belgrave and
K. Cho and
A. Oh},
 title = {FlashAttention: Fast and Memory-Efficient Exact Attention with IO-Awareness},
 url = {http://papers.nips.cc/paper\_files/paper/2022/hash/67d57c32e20fd0a7a302cb81d36e40d5-Abstract-Conference.html},
 year = {2022}
}

@inproceedings{Kwon2023Efficient,
author = {Kwon, Woosuk and Li, Zhuohan and Zhuang, Siyuan and Sheng, Ying and Zheng, Lianmin and Yu, Cody Hao and Gonzalez, Joseph and Zhang, Hao and Stoica, Ion},
title = {Efficient Memory Management for Large Language Model Serving with PagedAttention},
year = {2023},
isbn = {9798400702297},
publisher = {Association for Computing Machinery},
address = {New York, NY, USA},
url = {https://doi.org/10.1145/3600006.3613165},
doi = {10.1145/3600006.3613165},
booktitle = {Proceedings of the 29th Symposium on Operating Systems Principles},
pages = {611–626},
numpages = {16},
location = {Koblenz, Germany},
series = {SOSP '23}
}

@misc{ollama2024,
 author = {{Ollama Team}},
 howpublished = {\url{https://ollama.ai}},
 title = {Ollama: Open LLM Deployment Made Simple},
 year = {2024}
}

@inproceedings{muennighoff2023mteb,
 address = {Dubrovnik, Croatia},
 author = {Muennighoff, Niklas  and
Tazi, Nouamane  and
Magne, Loic  and
Reimers, Nils},
 booktitle = {Proceedings of the 17th Conference of the European Chapter of the Association for Computational Linguistics},
 doi = {10.18653/v1/2023.eacl-main.148},
 editor = {Vlachos, Andreas  and
Augenstein, Isabelle},
 pages = {2014--2037},
 publisher = {Association for Computational Linguistics},
 title = {{MTEB}: Massive Text Embedding Benchmark},
 url = {https://aclanthology.org/2023.eacl-main.148},
 year = {2023}
}

@inproceedings{liu2024llmembed,
    title = "{LLME}mbed: Rethinking Lightweight {LLM}{'}s Genuine Function in Text Classification",
    author = "Liu, Chun  and
      Zhang, Hongguang  and
      Zhao, Kainan  and
      Ju, Xinghai  and
      Yang, Lin",
    editor = "Ku, Lun-Wei  and
      Martins, Andre  and
      Srikumar, Vivek",
    booktitle = "Proceedings of the 62nd Annual Meeting of the Association for Computational Linguistics (Volume 1: Long Papers)",
    month = aug,
    year = "2024",
    address = "Bangkok, Thailand",
    publisher = "Association for Computational Linguistics",
    url = "https://aclanthology.org/2024.acl-long.433/",
    doi = "10.18653/v1/2024.acl-long.433",
    pages = "7994--8004",
}

@inproceedings{zeyu2025large,
  title={Large Reasoning Models Know How to Think Efficiently},
  author={Xing, Zeyu and Li, Xing and Zhen, Huiling and Yu, Xianzhi and Yuan, Mingxuan and Pan, Sinno Jialin},
  booktitle={ES-FoMo III: 3rd Workshop on Efficient Systems for Foundation Models},
  year={2025},
  url = "https://openreview.net/forum?id=pLKDeGm2t1"
}

@misc{cobbe2021gsm8k,
 author = {Karl Cobbe and Vineet Kosaraju and Mohammad Bavarian and Mark Chen and Heewoo Jun and Lukasz Kaiser and Matthias Plappert and Jerry Tworek and Jacob Hilton and Reiichiro Nakano and Christopher Hesse and John Schulman},
 journal = {ArXiv preprint},
 title = {Training Verifiers to Solve Math Word Problems},
 url = {https://arxiv.org/abs/2110.14168},
 volume = {abs/2110.14168},
 year = {2021}
}

@inproceedings{yao2023tree,
 author = {Shunyu Yao and
Dian Yu and
Jeffrey Zhao and
Izhak Shafran and
Tom Griffiths and
Yuan Cao and
Karthik Narasimhan},
 bibsource = {dblp computer science bibliography, https://dblp.org},
 booktitle = {Advances in Neural Information Processing Systems 36: Annual Conference
on Neural Information Processing Systems 2023, NeurIPS 2023, New Orleans,
LA, USA, December 10 - 16, 2023},
 editor = {Alice Oh and
Tristan Naumann and
Amir Globerson and
Kate Saenko and
Moritz Hardt and
Sergey Levine},
 title = {Tree of Thoughts: Deliberate Problem Solving with Large Language Models},
 url = {http://papers.nips.cc/paper\_files/paper/2023/hash/271db9922b8d1f4dd7aaef84ed5ac703-Abstract-Conference.html},
 year = {2023}
}

@inproceedings{lightman2023lets,
 author = {Hunter Lightman and
Vineet Kosaraju and
Yuri Burda and
Harrison Edwards and
Bowen Baker and
Teddy Lee and
Jan Leike and
John Schulman and
Ilya Sutskever and
Karl Cobbe},
 bibsource = {dblp computer science bibliography, https://dblp.org},
 booktitle = {The Twelfth International Conference on Learning Representations,
{ICLR} 2024, Vienna, Austria, May 7-11, 2024},
 publisher = {OpenReview.net},
 title = {Let's Verify Step by Step},
 url = {https://openreview.net/forum?id=v8L0pN6EOi},
 year = {2024}
}

@inproceedings{yuan2024coe,
 author = {Wang, Yiming and Zhang, Pei and Yang, Baosong and Wong, Derek and Wang, Rui},
 booktitle = {International Conference on Representation Learning},
 editor = {Y. Yue and A. Garg and N. Peng and F. Sha and R. Yu},
 pages = {70938--70970},
 title = {Latent Space Chain-of-Embedding Enables Output-free LLM Self-Evaluation},
 url = {https://proceedings.iclr.cc/paper_files/paper/2025/file/b0b1cfc8ede53f452cabf8b9cf4eef76-Paper-Conference.pdf},
 volume = {2025},
 year = {2025}
}

@inproceedings{hendrycks2021math,
 author = {Hendrycks, Dan and Burns, Collin and Kadavath, Saurav and Arora, Akul and Basart, Steven and Tang, Eric and Song, Dawn and Steinhardt, Jacob},
 booktitle = {Proceedings of the Neural Information Processing Systems Track on Datasets and Benchmarks},
 editor = {J. Vanschoren and S. Yeung},
 pages = {},
 title = {Measuring Mathematical Problem Solving With the MATH Dataset},
 url = {https://datasets-benchmarks-proceedings.neurips.cc/paper_files/paper/2021/file/be83ab3ecd0db773eb2dc1b0a17836a1-Paper-round2.pdf},
 volume = {1},
 year = {2021}
}

@inproceedings{chen2023theoremqa,
 address = {Singapore},
 author = {Chen, Wenhu  and
Yin, Ming  and
Ku, Max  and
Lu, Pan  and
Wan, Yixin  and
Ma, Xueguang  and
Xu, Jianyu  and
Wang, Xinyi  and
Xia, Tony},
 booktitle = {Proceedings of the 2023 Conference on Empirical Methods in Natural Language Processing},
 doi = {10.18653/v1/2023.emnlp-main.489},
 editor = {Bouamor, Houda  and
Pino, Juan  and
Bali, Kalika},
 pages = {7889--7901},
 publisher = {Association for Computational Linguistics},
 title = {{T}heorem{QA}: A Theorem-driven Question Answering Dataset},
 url = {https://aclanthology.org/2023.emnlp-main.489},
 year = {2023}
}

@inproceedings{chen2025inside,
 author = {Chao Chen and
Kai Liu and
Ze Chen and
Yi Gu and
Yue Wu and
Mingyuan Tao and
Zhihang Fu and
Jieping Ye},
 bibsource = {dblp computer science bibliography, https://dblp.org},
 booktitle = {The Twelfth International Conference on Learning Representations,
{ICLR} 2024, Vienna, Austria, May 7-11, 2024},
 publisher = {OpenReview.net},
 title = {{INSIDE:} LLMs' Internal States Retain the Power of Hallucination
Detection},
 url = {https://openreview.net/forum?id=Zj12nzlQbz},
 year = {2024}
}

@inproceedings{beigi2024internalinspector,
    title = "{I}nternal{I}nspector $I^2$: Robust Confidence Estimation in {LLM}s through Internal States",
    author = "Beigi, Mohammad  and
      Shen, Ying  and
      Yang, Runing  and
      Lin, Zihao  and
      Wang, Qifan  and
      Mohan, Ankith  and
      He, Jianfeng  and
      Jin, Ming  and
      Lu, Chang-Tien  and
      Huang, Lifu",
    editor = "Al-Onaizan, Yaser  and
      Bansal, Mohit  and
      Chen, Yun-Nung",
    booktitle = "Findings of the Association for Computational Linguistics: EMNLP 2024",
    month = nov,
    year = "2024",
    address = "Miami, Florida, USA",
    publisher = "Association for Computational Linguistics",
    url = "https://aclanthology.org/2024.findings-emnlp.751/",
    doi = "10.18653/v1/2024.findings-emnlp.751",
    pages = "12847--12865"
}

@inproceedings{
chen2025probing_right,
title={Reasoning Models Know When They{\textquoteright}re Right: Probing Hidden States for Self-Verification},
author={Anqi Zhang and Yulin Chen and Jane Pan and Chen Zhao and Aurojit Panda and Jinyang Li and He He},
booktitle={Second Conference on Language Modeling},
year={2025},
url={https://openreview.net/forum?id=O6I0Av7683}
}

@misc{zhang2025asrr,
 author = {Xiaoyun Zhang and Jingqing Ruan and Xing Ma and Yawen Zhu and Haodong Zhao and Hao Li and Jiansong Chen and Ke Zeng and Xunliang Cai},
 journal = {ArXiv preprint},
 title = {When to Continue Thinking: Adaptive Thinking Mode Switching for Efficient Reasoning},
 url = {https://arxiv.org/abs/2505.15400},
 volume = {abs/2505.15400},
 year = {2025}
}

@misc{wang2025pats,
 author = {Yi Wang and Junxiao Liu and Shimao Zhang and Jiajun Chen and Shujian Huang},
 journal = {ArXiv preprint},
 title = {PATS: Process-Level Adaptive Thinking Mode Switching},
 url = {https://arxiv.org/abs/2505.19250},
 volume = {abs/2505.19250},
 year = {2025}
}

@inproceedings{wang2025dots,
 author = {Yue, Murong and Yao, Wenlin and Mi, Haitao and Yu, Dian and Yao, Ziyu and Yu, Dong},
 booktitle = {International Conference on Representation Learning},
 editor = {Y. Yue and A. Garg and N. Peng and F. Sha and R. Yu},
 pages = {37976--37997},
 title = {DOTS: Learning to Reason Dynamically in LLMs via Optimal Reasoning Trajectories Search},
 url = {https://proceedings.iclr.cc/paper_files/paper/2025/file/5e5d6f9ac33ba9349ba7b2be9f21bad9-Paper-Conference.pdf},
 volume = {2025},
 year = {2025}
}

@misc{belitsky2025kvsteer,
 author = {Max Belitsky and Dawid J. Kopiczko and Michael Dorkenwald and M. Jehanzeb Mirza and Cees G. M. Snoek and Yuki M. Asano},
 journal = {ArXiv preprint},
 title = {KV Cache Steering for Inducing Reasoning in Small Language Models},
 url = {https://arxiv.org/abs/2507.08799},
 volume = {abs/2507.08799},
 year = {2025}
}

@misc{guo2025deepseek,
 author = {DeepSeek-AI},
 journal = {ArXiv preprint},
 title = {DeepSeek-R1: Incentivizing Reasoning Capability in LLMs via Reinforcement Learning},
 url = {https://arxiv.org/abs/2501.12948},
 volume = {abs/2501.12948},
 year = {2025}
}

@misc{yang2025qwen3,
 author = {QwenTeam, Alibaba Group},
 journal = {ArXiv preprint},
 title = {Qwen3 Technical Report},
 url = {https://arxiv.org/abs/2505.09388},
 volume = {abs/2505.09388},
 year = {2025}
}

@misc{math5002025,
 author = {{OpenAI / HuggingFaceH4 / Vals AI}},
 howpublished = {HuggingFace / Vals AI Benchmark / Datasets},
 note = {Derived from “Measuring Mathematical Problem Solving With the MATH Dataset”; subset of 500 test problems},
 title = {MATH-500: A 500-Problem Subset of the MATH Benchmark},
 url = {https://huggingface.co/datasets/HuggingFaceH4/MATH-500},
 year = {2025}
}

@misc{grattafiori2024llama3,
 author = {LlamaTeam,AI@Meta},
 journal = {ArXiv preprint},
 title = {The Llama 3 Herd of Models},
 url = {https://arxiv.org/abs/2407.21783},
 volume = {abs/2407.21783},
 year = {2024}
}

@misc{yang2025qwen2,
 author = {QwenTeam, Alibaba Group},
 journal = {ArXiv preprint},
 title = {Qwen2 Technical Report},
 url = {https://arxiv.org/abs/2407.10671},
 volume = {abs/2407.10671},
 year = {2024}
}

@inproceedings{yangattentionpredictor,
  title={AttentionPredictor: Temporal Patterns Matter for KV Cache Compression},
  author={Yang, Qingyue and Wang, Jie and Li, Xing and Wang, Zhihai and Chen, Chen and Chen, Lei and Yu, Xianzhi and Liu, Wulong and Hao, Jianye and Yuan, Mingxuan and others},
  booktitle={The Thirty-ninth Annual Conference on Neural Information Processing Systems}
}

@article{li2025kvtuner,
  title={KVTuner: Sensitivity-Aware Layer-Wise Mixed-Precision KV Cache Quantization for Efficient and Nearly Lossless LLM Inference},
  author={Li, Xing and Xing, Zeyu and Li, Yiming and Qu, Linping and Zhen, Hui-Ling and Liu, Wulong and Yao, Yiwu and Pan, Sinno Jialin and Yuan, Mingxuan},
  journal={arXiv preprint arXiv:2502.04420},
  year={2025}
}
\bibliographystyle{iclr2026_conference}

\newpage
\appendix
\section{The Use of Large Language Models (LLMs)}
We acknowledge the use of a Large Language Model (LLM) to support the preparation of this 
manuscript. The LLM was employed exclusively for editorial purposes, such as refining the 
clarity of exposition, improving grammar and readability, and polishing the overall presentation. 
At times, it was also used to suggest alternative phrasings for technical descriptions in order 
to make the arguments more accessible. 

Importantly, the LLM did not contribute to the conceptual development, methodology, or 
experimental design of this work. All ideas, including the proposal to treat the KV cache as 
a reusable representation, the development of KV-CoE for output-free self-evaluation, and the 
design of KV-based adaptive Fast/Slow Thinking Switching for token-efficient reasoning, were 
conceived and implemented solely by the authors. The LLM was not used to generate research 
results, proofs, or derivations. 

All scientific claims, analyses, and conclusions presented in this paper remain the full 
responsibility of the authors. Any text initially produced with LLM assistance was carefully 
reviewed, revised, and verified prior to inclusion.

\section{CoE Usage, KV-CoE Design, and Token vs.\ Layer Dimension}
\label{appendix:coe}

\paragraph{CoE usage in reasoning benchmarks.}
In both the original CoE framework and our work, CoE is used as a \emph{single-sample confidence estimator}. Given one completed reasoning trace, CoE outputs a scalar confidence score, and is evaluated via AUROC/FPR95/AUPR. CoE is \emph{not} used for pass@k reranking, and it does not modify decoding or generation.

\paragraph{KV cache vs.\ hidden states across the layer dimension.}
The KV cache stores attention key/value tensors (per layer/head/token) and does not provide a drop-in substitute for full hidden states along the layer dimension. In particular, directly replacing hidden states with KV cache within the original layer-wise CoE design yields poor performance. This reflects a structural mismatch: the KV cache is inherently token-centric and head-structured, whereas hidden-state CoE explicitly tracks layer-wise representations.

\paragraph{Ablation: layer-wise aggregation with KV-CoE.}
To make this limitation explicit, we evaluate a layer-wise variant of KV-CoE on \texttt{meta-llama/Llama-3.1-8B-Instruct}. Results are shown in Tables~\ref{tab:kvcoe-layer-math} and~\ref{tab:kvcoe-layer-theoremqa}. Layer-wise KV-CoE performs substantially worse than our token-centric KV-CoE, confirming that KV cache should not be treated as a layer-wise embedding.

\begin{table}[htbp]
\centering
\begin{tabular}{lccc}
\toprule
Method & AUROC $\uparrow$ & FPR95 $\downarrow$ & AUPR $\uparrow$ \\
\midrule
KV-CoE-R (layer) & 62.69 & 82.28 & 45.43 \\
KV-CoE-C (layer) & 64.09 & 81.16 & 47.78 \\
\bottomrule
\end{tabular}
\caption{Layer-wise aggregation ablation of KV-CoE on \textbf{MATH} using \texttt{meta-llama/Llama-3.1-8B-Instruct}.}
\label{tab:kvcoe-layer-math}
\end{table}

\begin{table}[htbp]
\centering
\begin{tabular}{lccc}
\toprule
Method & AUROC $\uparrow$ & FPR95 $\downarrow$ & AUPR $\uparrow$ \\
\midrule
KV-CoE-R (layer) & 47.78 & 94.24 & 17.70 \\
KV-CoE-C (layer) & 48.61 & 92.68 & 18.02 \\
\bottomrule
\end{tabular}
\caption{Layer-wise aggregation ablation of KV-CoE on \textbf{TheoremQA} using \texttt{meta-llama/Llama-3.1-8B-Instruct}.}
\label{tab:kvcoe-layer-theoremqa}
\end{table}

\paragraph{Design implication: token-centric KV-CoE.}
Motivated by the above, we redesign KV-CoE to operate along the \emph{token dimension}, using token-wise aggregation that matches the semantics of the KV cache. Concretely, KV-CoE forms per-token embeddings from value vectors and aggregates them over token positions (and optionally across heads/layers) to produce the trace-level confidence score.

\paragraph{Pooling strategies summary.}
For clarity, Table~\ref{tab:pooling-summary} summarizes the pooling/aggregation strategies used in this paper (see also L320--322 and L455--456).

\begin{table}[htbp]
\centering
\resizebox{\linewidth}{!}{%
\begin{tabular}{lllll}
\toprule
Task & Source & Head Agg. & Position Agg. & Layer Agg. \\
\midrule
KV-CoE & Value vectors from KV cache & Concatenate & Per-token embedding & Average \\
Fast/Slow Thinking & Key + Value & Concatenate & Sum over selected tokens & Average (no normalization) \\
\bottomrule
\end{tabular}%
}
\caption{Summary of pooling strategies used in this work.}
\label{tab:pooling-summary}
\end{table}

\section{Hidden-State and Chance Baselines on MTEB Classification Subsets}
\label{appendix:mteb}

To contextualize the MTEB classification results in Table~\ref{tab:kv_mteb}, we report two additional baselines evaluated under the \emph{same} downstream protocol (identical pooling, projection, and classifier setup as used for KV-cache embeddings).

\paragraph{Hidden-state embeddings.}
We extract hidden-state embeddings from \texttt{meta-llama/Llama-3.1-8B-Instruct} and apply the same aggregation pipeline used for KV-cache embeddings. Table~\ref{tab:hidden-mteb} shows that hidden-state embeddings achieve accuracies close to those of KV-cache embeddings under identical pooling and projection, suggesting that the observed gap to dedicated embedding models is dominated by the downstream protocol rather than the choice between hidden states vs.\ KV caches.

\begin{table}[htbp]
\centering
\begin{tabular}{l c}
\toprule
Dataset & Accuracy (Hidden State) \\
\midrule
AmazonCounterfactual & 0.3530 \\
DBpedia & 0.5937 \\
FinancialPhrasebank & 0.6254 \\
TweetTopic & 0.3714 \\
\bottomrule
\end{tabular}
\caption{Accuracy of hidden-state embeddings on selected MTEB classification tasks under the same pooling/projection pipeline as KV-cache embeddings.}
\label{tab:hidden-mteb}
\end{table}

\paragraph{Random-embedding baseline.}
We additionally include a random-embedding baseline to sanity-check that the evaluation pipeline is not degenerate. Concretely, we replace the model-derived embedding with an i.i.d.\ random vector $e_{\text{rand}}\in\mathbb{R}^d$ (matched to the embedding dimension $d$ used in our pipeline), while keeping the \emph{same} pooling/projection interface (where applicable) and the same downstream classifier and evaluation protocol. Table~\ref{tab:random-mteb} reports the resulting accuracy. The consistent margin between random embeddings and hidden/KV embeddings—particularly on DBpedia and FinancialPhrasebank—indicates that the pipeline is functional and that model-derived embeddings capture non-trivial task structure beyond what is achievable with unstructured random features.

\begin{table}[htbp]
\centering
\begin{tabular}{l c}
\toprule
Dataset & Accuracy (Random) \\
\midrule
AmazonCounterfactualClassification & 0.5224 \\
DBpediaClassification & 0.0716 \\
FinancialPhrasebankClassification & 0.3274 \\
TweetTopicClassification & 0.1591 \\
\bottomrule
\end{tabular}
\caption{MTEB classification performance when representations are replaced by a random baseline embedding $e \in \mathbb{R}^{256}$, whose entries are drawn i.i.d.\ from a standard normal distribution.}
\label{tab:random-mteb}
\end{table}

\paragraph{Scope.}
These MTEB subsets are reported only to illustrate that KV-derived embeddings are \emph{cost-efficiently comparable} under our protocol; we do not claim they are superior to dedicated embedding models, nor that they are reliable for global semantic retrieval across diverse domains.

\section{Difficulty Labels and Prompt Control for Fast/Slow Thinking Switching}
\label{appendix:fs-labels-prompts}

\paragraph{Difficulty labels for training the switching model.}
We assign each instance a discrete difficulty label based on the correctness of \emph{fast} vs.\ \emph{slow} thinking outputs, using the following four-level scheme:

\begin{table}[htbp]
\centering
\begin{tabular}{clc}
\toprule
Difficulty & Description & Count \\
\midrule
0   & Both fast and slow thinking are correct (very easy) & 286 \\
25  & Both are correct but require longer generation & 3{,}467 \\
75  & Only slow thinking is correct & 860 \\
100 & Neither is correct & 2{,}887 \\
\midrule
\multicolumn{2}{l}{Total} & 7{,}500 \\
\bottomrule
\end{tabular}%
\caption{Difficulty labels used to train the fast/slow switching estimator.}
\label{tab:fs-difficulty-labels}
\end{table}

\paragraph{Fast vs.\ slow thinking prompts.}
We use a Large Reasoning Model (LRM) that supports explicit reasoning via \texttt{<think>} blocks. Fast/slow thinking is controlled \emph{purely by prompt}, not by KV cache. Specifically, after the user prompt:
(i) \textbf{Fast thinking} inserts an empty thinking block \texttt{<think>\textbackslash n</think>};
(ii) \textbf{Slow thinking} inserts an open thinking block \texttt{<think>\textbackslash n}, which encourages step-by-step reasoning.
Model-specific special tokens (e.g., \texttt{<|im\_start|>}, \texttt{<|im\_end|>}) are adapted per backbone; the KV-based controller only chooses between the fast and slow prompt variants.

\section{Hyper-parameter Selection for KVClassifier}
\label{appendix:kv_hyperparam}

To better understand the effect of hyper-parameters on KV-based classification, 
we conduct a systematic study by varying the number of layers pooled and the number of 
tokens selected from the end of the sequence. Importantly, we fix the \emph{total amount of KV data} 
to be approximately constant across configurations (256 token $\times$ layer units). 
This ensures a fair comparison: for example, selecting 8 layers with 32 tokens, 
4 layers with 64 tokens, or 2 layers with 128 tokens all yield the same KV budget.  

\begin{table}[htbp]
\centering
\resizebox{0.8\textwidth}{!}{
\begin{tabular}{llcccc}
\toprule
\textbf{Model} & \textbf{Dataset} & \textbf{Method} & \textbf{8L, Len=32} & \textbf{4L, Len=64} & \textbf{2L, Len=128} \\
\midrule
\multirow{4}{*}{DeepSeek-14B}
 & \multirow{2}{*}{GSM8K} 
   & KV-Classification & 0.845 / 218 & 0.845 / 218 & \textbf{0.845} / 218\; \\
 &  & KV-Generative & \textbf{0.835} / 242 & 0.825 / 232 & 0.805 / 217\; \\
 & \multirow{2}{*}{MATH500}
   & KV-Classification & 0.536 / 540 & 0.550 / 905 & \textbf{0.578} / 1506 \\
 &  & KV-Generative & 0.538 / 524 & 0.550 / 544 & \textbf{0.566} / 657\; \\
\midrule
\multirow{4}{*}{Qwen3-8B}
 & \multirow{2}{*}{GSM8K}
   & KV-Classification & 0.904 / 211 & 0.904 / 217\; & \textbf{0.914} / 554\; \\
 &  & KV-Generative & \textbf{0.892} / 273 & 0.886 / 276\; & 0.881 / 257\; \\
 & \multirow{2}{*}{MATH500}
   & KV-Classification & 0.570 / 736 & 0.598 / 3673 & \textbf{0.604} / 3963 \\
 &  & KV-Generative & \textbf{0.578} / 727 & 0.524 / 933\; & 0.550 / 837\; \\
\bottomrule
\end{tabular}}
\vspace{-0.2cm}
\caption{Hyper-parameter selection results for KV-Classification and KV-Generative. 
Values are reported as Accuracy / \#Tokens. Best accuracy for each dataset–method pair is in bold.}
\label{tab:hyperparam}
\vspace{-0.5cm}
\end{table}

Table~\ref{tab:hyperparam} summarizes the results on GSM8K and MATH500 for both DeepSeek-R1-14B 
and Qwen3-8B, under KV-Classification and KV-Generative settings.  
We observe that while performance varies slightly with the allocation of layer vs.\ token depth, 
the overall trends are consistent: (i) accuracy remains competitive across different allocations, 
and (ii) increasing token coverage (e.g., 2L $\times$ 128) tends to favor more complex datasets 
such as MATH500, whereas shallow but wider layer coverage (e.g., 8L $\times$ 32) can suffice 
for GSM8K.

\end{document}